\documentclass[letterpaper, 10 pt, conference]{ieeeconf}  

\IEEEoverridecommandlockouts                              

\usepackage{graphics} 
\usepackage{epsfig} 
\usepackage{mathptmx} 
\usepackage{times} 
\usepackage{amsmath} 
\usepackage{amssymb}  
\usepackage{booktabs}
\usepackage{caption}
\usepackage{graphicx}
\usepackage{float} 
\usepackage{subfigure}
\usepackage{subcaption}

\title{\LARGE \bf
Goal State Generation for Robotic Manipulation Based on Linguistically Guided Hybrid Gaussian Diffusion
}

\author{Yichen Xu, Faliang Chang, Chunsheng Liu and Dexin Wang
}

\begin{document}

\maketitle
\thispagestyle{empty}
\pagestyle{empty}

\begin{abstract}

In robotic manipulation tasks, achieving a designated target state for the manipulated object is often essential to facilitate motion planning for robotic arms. Specifically, in tasks such as hanging a mug, the mug must be positioned within a feasible region around the hook. Previous approaches have enabled the generation of multiple feasible target states for mugs; however, these target states are typically generated randomly, lacking control over the specific generation locations. This limitation makes such methods less effective in scenarios where constraints exist, such as hooks already occupied by other mugs or when specific operational objectives must be met. Moreover, due to the frequent physical interactions between the mug and the rack in real-world hanging scenarios, imprecisely generated target states from end-to-end models often result in overlapping point clouds. This overlap adversely impacts subsequent motion planning for the robotic arm. To address these challenges, we propose a Linguistically Guided Hybrid Gaussian Diffusion (LHGD) network for generating manipulation target states, combined with a gravity coverage coefficient-based method for target state refinement. To evaluate our approach under a language-specified distribution setting, we collected multiple feasible target states for 10 types of mugs across 5 different racks with 10 distinct hooks. Additionally, we prepared five unseen mug designs for validation purposes. Experimental results demonstrate that our method achieves the highest success rates across single-mode, multi-mode, and language-specified distribution manipulation tasks. Furthermore, it significantly reduces point cloud overlap, directly producing collision-free target states and eliminating the need for additional obstacle avoidance operations by the robotic arm.

\end{abstract}

\section{INTRODUCTION}

Robotic manipulation tasks have long been one of the most challenging problems in the field of robotics. These tasks involve various sub-tasks such as grasping, placing, and assembling, all of which require the robot to move objects to a target state. Many current methods [1,2,3,4,5] teach robots complex tasks through human demonstrations. However, such approaches require a large amount of demonstration data to generalize well to novel initial states. Therefore, if a robot can infer the object’s state at the completion of a task from any given initial state, it can overcome the dependency and sensitivity to the initial state, significantly enhancing the system's stability and robustness.

Some more complex robotic manipulation tasks require the robot to place an object in a specific location relative to another object, with both objects typically in contact (e.g., hanging a mug on a rack). To complete these tasks, the robot needs to predict the target state of the object based on scene information such as object poses and point clouds. This state needs to be highly accurate to avoid issues like collisions or overlaps between objects, which can affect downstream path planning and execution. Previous studies have been able to generate multiple reasonable target states, but both diffusion models [6] and generative models like CVAE [7] generate results randomly, making it difficult to control the exact target state. This poses challenges for tasks requiring specific placement locations or in cases where some valid positions are occupied (Figure 1 left), making it difficult to complete placement tasks effectively. Moreover, given the high precision required for the object’s target state, collisions or overlaps between objects are unavoidable in predictions, which can hinder the successful completion of the task (Figure 1 right).

   \begin{figure}[thpb]
      \centering
      \includegraphics[width=\columnwidth]{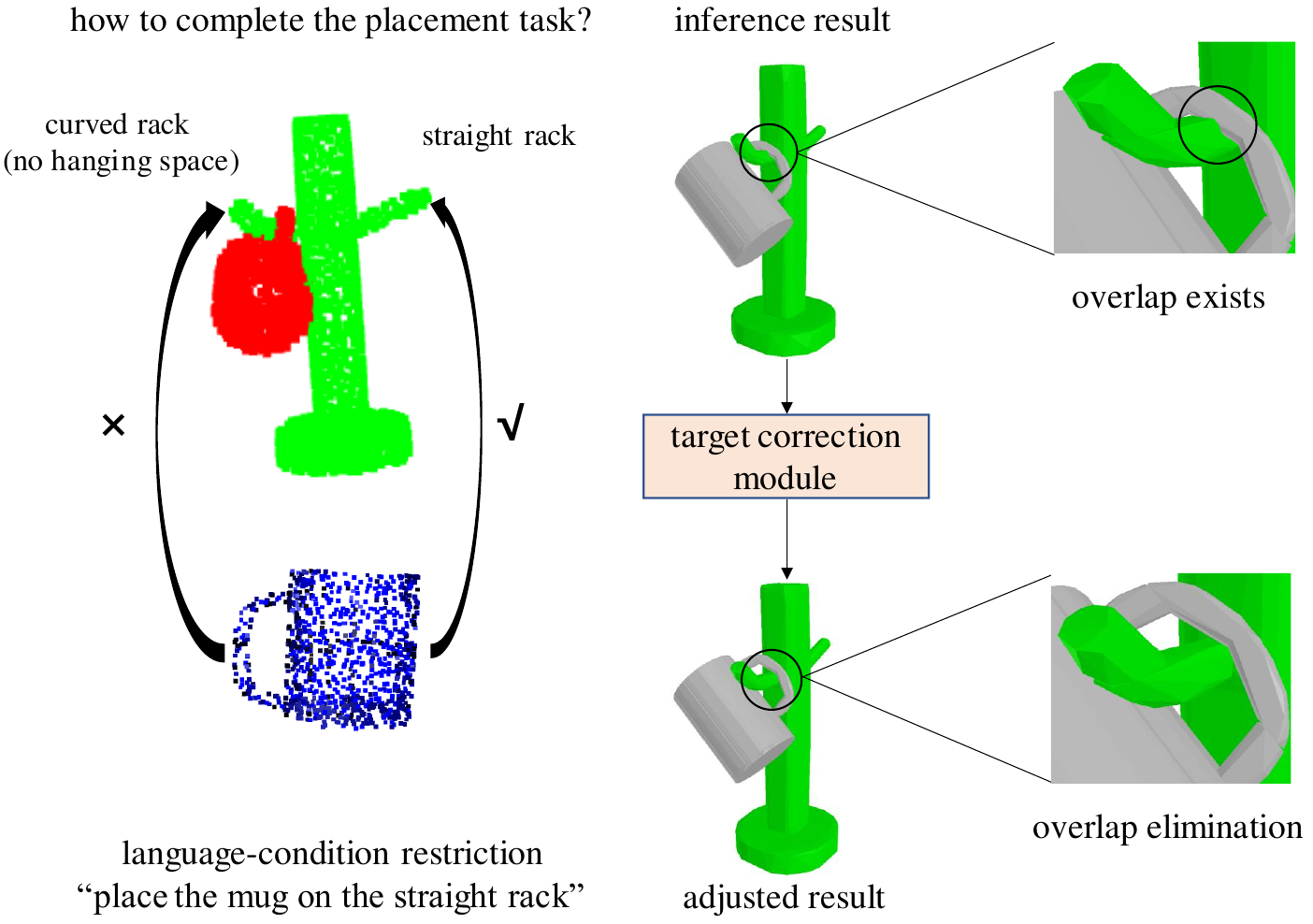}
      \caption{Diagram of generation controlled by linguistic conditions (left) and illustration of de-overlapping module (right)}
      \label{figurelabel}
   \end{figure}

To address these issues, we propose a Linguistic Hybrid Gaussian Diffusion (LHGD) network for generating operation target states conditioned on language. By collecting data from five types of mugs, five types of racks, and ten different hanging positions (i.e., ten different linguistic conditions), we construct a custom dataset and use control statements, point clouds of mugs and racks, and other information to control the generated target state through linguistic conditions.

Our contributions are as follows:

\begin{itemize}

\item To address the challenge of directly adding Gaussian noise to pose information for diffusion, we propose the LHGD operation target state generation network, which decouples diffusion generation for rotation and translation based on isotropic Gaussian distribution and normal distribution, respectively. This approach enhances the stability of the generated target rigid body point clouds while increasing the diversity of the generated target states.
\item To solve the issue of collision in generated target states, we introduce a post-processing method based on the gravity drop coverage coefficient to reduce overlap. A post-processing module built on the DDPM diffusion model fine-tunes the predicted target poses, further improving the model's inference accuracy. This significantly reduces overlap and collision between objects, allowing the robot to use collision-free poses directly during manipulation, without requiring additional obstacle avoidance actions.

\end{itemize}

\section{RELATED WORK}

\textbf{Target State Generation.}
In robotic manipulation tasks, generating a target state for the object to be manipulated is crucial for subsequent path planning to complete the task. Previous research in the field of operation target state generation [8, 9, 10, 11] has explored various approaches, where robots learn transformations from an initial state to a target state through a small number of human demonstrations. While these methods improve the generalization capability compared to earlier demonstration-based techniques, they perform poorly in tasks involving multimodal distributions (multiple possible target states). Our approach addresses this limitation by leveraging diffusion models, which effectively capture and generate multiple modes within multimodal distributions, enabling better performance in tasks with multiple feasible solutions.TAX-PoseD [12] leverages a CVAE network to accomplish multi-distribution tasks but lacks the ability to specify the generated target states. In contrast, our method integrates language conditions into multi-distribution tasks, allowing the input statements to control the precise location of the generated target states.

\textbf{Diffusion Models.}
Diffusion models are a class of generative models, where the core idea is to progressively degrade data by adding noise, and then learn how to reverse this degradation process to recover the original structure or generate new, realistic data points. Recently, diffusion models have been widely applied in the field of robotic manipulation [13, 14, 15], achieving promising results. Some research has also combined diffusion models with robotic operation target state generation. However, methods such as RPDiff [16] and StructDiffusion [17] have their limitations. The former cannot control the specific target states generated under multimodal distributions, while the latter targets desktop rearrangement tasks with overly simple constraints on the target state, resulting in lower accuracy requirements. In contrast, our approach uses linguistic conditions to control the target state of a mug hanging task. It can handle specified mode distribution tasks while also satisfying the strong constraints between the mug and the rack, ensuring high accuracy in the generated target states.

\section{Problem Statement and Assumptions}

In robotic manipulation tasks, the core of generating an operational target state lies in determining a precise and feasible target state that a manipulated object should reach to complete the task. To accommodate a wide variety of manipulated objects—including rigid bodies, soft bodies, and articulated objects—using the object’s point cloud to represent the target state is an ideal approach. This representation provides rich geometric and spatial information in three-dimensional space, enabling the generated target state to encapsulate detailed surface features and contact regions of the object. Let $P_i$ and $P_t$ denote the point clouds of the manipulated object in its initial and target states, respectively, and $P_e$ represent the point cloud of the operational environment. We describe the target state generation task as a mapping relationship:
\begin{equation}
P_t=f\left(P_i, P_e, C\right)
\end{equation}
where $f$ denotes the mapping to complete the task, and $C$ represents various condition information specific to the task (e.g. language, time steps, etc.).

Specifically, when the manipulated object is a non-deformable rigid body, we aim to avoid deformation in the predicted point cloud that could interfere with subsequent tasks. Thus, for rigid bodies, we specialize the task of generating the target operational state as target pose generation, using this target pose to map the object's point cloud to the desired state. This approach helps prevent any unintended deformation in the predicted point cloud. Similarly, we denote the initial and target poses of the object $T_i$ as and $T_t$ , respectively, and describe the generation of a rigid body's target state as follows:
\begin{equation}
T_t=f\left(T_i, P_i, P_e, C\right), \quad P_t=p\left(T_t\right)
\end{equation}
where $p$ denotes the mapping from the object's target pose to the target point cloud.

Although it is possible to obtain a predicted target state for the manipulated object, the relational constraints between the manipulated object and the environment are also crucial for robotic manipulation tasks. For complex tasks (e.g. hanging a mug), it is not only necessary for the generated target state to lie near the distribution of true values, but it must also avoid collisions with the environment—especially with other objects that have special constraints relative to the manipulated object in the task. This ensures seamless path planning and execution for the robotic arm. Since both the generated target state and the true value exist as distributions, we denote them by and , respectively. We propose that the task of target state generation is accomplished when the following inequalities are satisfied:
\begin{equation}
\left\{\begin{array}{l}
D_{\mathit{KL}}\left(p\left(T_{\mathit{t}}\right) \| p\left(T_{\mathit{g}}\right)\right) \leq \epsilon_{\mathit{KL}}, \\
M_{\mathit{t}} \cap M_{\mathit{e}} = \emptyset .
\end{array}\right.
\end{equation}
where $D_{\mathit{KL}}$ represents the Kullback-Leibler divergence between the two distributions, and $\epsilon_{\mathit{KL}}$ denotes the acceptable error threshold. $M_{\mathit{t}}$ and $M_{\mathit{e}}$ represent the models of the manipulated object and the environmental objects, respectively. This set of inequalities defines the completion conditions for target state generation: (1) the KL divergence between the generated target state and the true distribution is within the allowable range, and (2) there is no collision between the manipulated object and other environmental objects.

\section{LHGD FOMULATION}

\subsection{Hybrid Gaussian Diffusion Generation}

To address the deformation issue that arises when adding Gaussian noise to transformation matrices, we have improved standard Gaussian noise by using a mixture of Gaussian noise for both the noising and denoising of transformation matrices. The distribution of the mixed Gaussian noise consists of two components: a normal distribution and an isotropic Gaussian distribution [18], denoted as $\epsilon \sim N(0, I)$ and $g\sim \mathcal{I} \mathcal{G}_{S O(3)}\left(\mu,\epsilon^2\right)$.Specifically, the $\mathcal{I} \mathcal{G}$ distribution can be parameterized in the axis-angle form, where the axis is uniformly sampled, and the rotation angle $\omega \in[0, \pi]$ varies according to the density distribution:
\begin{equation}
f(\omega)=\frac{1-\cos \omega}{\pi} \sum_{l=0}^{\infty}(2 l+1) e^{-l(l+1) \epsilon^2} \frac{\sin \left(\left(l+\frac{1}{2}\right) \omega\right)}{\sin (\omega / 2)} .
\end{equation}

\textbf{Diffusion Process.}
The diffusion process is the forward procedure in a diffusion model, which gradually adds Gaussian noise to the original data (here, the ground-truth poses) until it ultimately becomes pure noise. This process consists of a series of time steps $t=1,2, \cdots, T$ , where $T$ represents the total number of steps in the diffusion process. The diffusion process can be viewed as an inhomogeneous discrete Markov chain, with the distribution at step $t$ represented as:
\begin{equation}
q\left(x_t \mid x_0\right)=N\left(x_t ; \sqrt{\bar{\alpha}_t} x_0,\left(1-\bar{\alpha}_t\right) I\right)
\end{equation}
where $x_0$ denotes the given initial data, $x_t$ represents the data at the $t$-th noise addition step, $\bar{\alpha}_{t}=\prod_{s=0}^{t} a_{s}$ and $a_{s}$ is the coefficient that controls the degree of mixture between the data and noise. For the diffusion process of the isotropic Gaussian distribution, we first need to define the logarithm of the rotation matrix:
\begin{equation}
\log R=\frac{\theta}{2 \sin \theta}\left(R^{T}-R\right)
\end{equation}
the equation allows us to scale the rotation matrix by converting it into a value in the Lie algebra. By analogy with the addition in the Euclidean diffusion model, the rotation in $SO$(3) is composed using matrix multiplication:
\begin{equation}
\lambda(\gamma,R)=\exp(\gamma\log(R))
\end{equation}
where the function $\lambda$ is the geodesic flow from the identity rotation matrix $I$ to $R$, with a value of $\gamma$. Therefore, for the isotropic Gaussian distribution, the diffusion process becomes:
\begin{equation}
q(R_t|R_0)=IG_{SO(3)}\left(\lambda(\sqrt{\bar{\alpha}_t},R_0),(1-\bar{\alpha}_t)\right)
\end{equation}
where $R_t$ is the rotation matrix at step $t$, and $R_0$ is the given initial data. The meaning of $\bar{\alpha}_t$ here is consistent with its role in the previous normal distribution.

\textbf{Reverse Process.}
The reverse process is a key step in diffusion models, where the principle is to gradually remove noise from the data to recover data that closely resembles the original, true form. Unlike the forward diffusion process, which incrementally adds noise, reverse process simulates each denoising step by learning a probabilistic model. Through iterative noise removal, this process generates samples with high fidelity and diversity. Similarly, the denoising process for Gaussian distributions can be described as:
\begin{equation}
p_\theta(x_{t-1}\mid x_t)=N(x_{t-1};\mu_\theta(x_t,t),\Sigma_\theta(x_t,t))
\end{equation}
where $\mu_\theta(x_t,t)$ represents the expected value after denoising, and $\Sigma_\theta(x_t,t)$ denotes the variance for each sampling step.For the denoising process of the isotropic Gaussian distribution, it can be described as:
\begin{equation}
p(x_{t-1}|x_t,x_0)=\mathcal{IG}_{SO(3)}(\tilde{\mu}(x_tx_0),\tilde{\beta}_t)
\end{equation}
where $\tilde{\mu}(x_tx_0)$ represents the denoised expectation, and we have:
\begin{equation}
\small
\tilde{\mu}(x_tx_0)=\lambda\left(\frac{\sqrt{\bar{\alpha}_{t-1}}\beta_t}{1-\bar{\alpha}_t},x_0\right)\lambda\left(\frac{\sqrt{\alpha_{t-1}}(1-\bar{\alpha}_{t-1})}{1-\bar{\alpha}_t},x_t\right)
\end{equation}

\section{LHGD Key Design}

\subsection{Network Design}

We propose a language-conditioned target state generation network based on hybrid Gaussian diffusion, named Linguistically Guided Hybrid Gaussian Diffusion (LHGD). This network not only performs single-mode and multi-mode generation tasks but also enables control over multiple randomly generated target poses through language conditioning. The overall framework of the network is illustrated in Figure 2. Our approach uses a diffusion model as the backbone framework, where the ground-truth poses in the dataset are iteratively noised through the diffusion process, resulting in random poses in space to simulate the diverse initial states of objects in the testing scenario. The network is trained to predict the added noise, and finally, the reverse diffusion process incrementally denoises the initial state to yield the predicted target state. However, unlike the diffusion process in fields such as image generation, where noise is added directly, the rotational matrix in pose data has intrinsic constraints among its elements. Directly adding Gaussian noise to it would lead to deformations in the object's point cloud or model. To address this, we propose a pose-decoupling diffusion module based on hybrid Gaussian noise, which performs decoupled diffusion on translation vectors and rotation matrices based on normal and isotropic Gaussian distributions, respectively. The following sections provide a detailed explanation of our method.

\begin{figure*}[t!]
	\centering
	\includegraphics[width=0.9\textwidth]{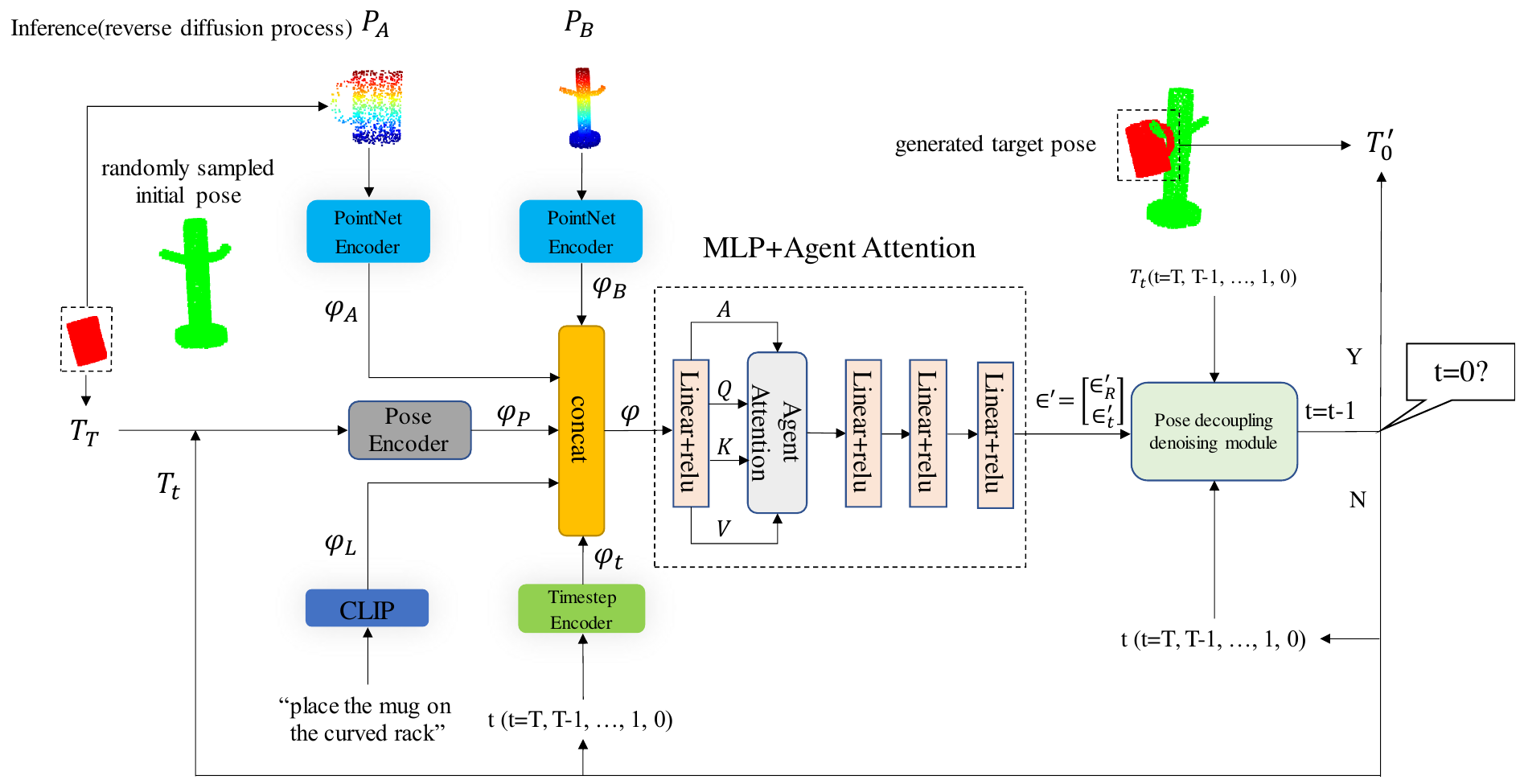} %
	\caption{Model Overview(reverse diffusion phase). During the training (diffusion) phase, we obtain the initial pose by adding noise to the target pose. An encoder is used to encode point cloud, pose, language, and timestep information into the feature space, which is then fused to obtain the complete feature information corresponding to the task in the scene. This information is then fed into the MLP structure combined with Agent Attention to predict the added noise. During the inference (reverse diffusion) phase, the trained parameters are used to predict the noise added to any initial mug pose, gradually restoring the target pose.}
	\label{figurelabel}
\end{figure*}

\textbf{Pose Encoder:}
Following the pose encoding approach outlined in LGrasp6D [19], we map the input pose information at different time steps $t$ into a high-dimensional feature space using a multi-layer perceptron (MLP). Given the high precision required for pose predictions in our task, encoding the pose facilitates the network's ability to learn the relationships between translation and rotation more effectively than directly inputting the raw pose information. We denote the resulting feature representation as $\varphi_P$.

\textbf{Point Cloud Encoder:}
After obtaining the point clouds $P_A$ and $P_B$ of the mug and the mug racks, respectively, we extract their features by training a point cloud encoder to capture latent representations. We select the PointNet [20] encoder for point feature extraction, as it has demonstrated strong performance in tasks such as point cloud segmentation and classification. Two separate encoders are employed for $P_A$ and $P_B$, with each encoder using an MLP network to map the position coordinates of each point to a higher-dimensional feature space. All points share the same network weights, ensuring the network’s invariance to point ordering. We denote the resulting feature representations as $\varphi_A$ and $\varphi_B$, respectively.

\textbf{Language Condition Encoder:}
Language is used as a conditional input to control the generation of target states. For multimodal distributions, we employ a unified statement: “place the mug on the arbitrary rack”, to indicate any arbitrary hook on the rack. For specified distributions, we use ten distinct statements: “place the mug on the longer/shorter/higher/lower/horizontal/tilted/rectangular/
cylindrical/curved/straight rack”, to correspond to ten unique hanging positions. To map these statements into the feature space, we utilize a pretrained CLIP model [21] to encode the ten statements. CLIP, trained on a diverse set of internet text, captures a wide semantic range and can effectively generate text embeddings with nuanced semantic distinctions. Given the large parameter size of the CLIP model, to avoid unnecessary computational overhead during training, we precompute the language features using CLIP during data collection and store them in the dataset as $\varphi_L$. This step is crucial for transforming the distributed relative placement task into a unimodal output.

\textbf{Time Step Encoder:}
In the diffusion process, noise is added randomly at each time step $t=1,2,\cdots,T$ resulting in each noisy pose being associated with a specific time step $t$. To facilitate model recognition and processing of temporal information in sequential data, we encode $t$, expanding the input feature dimensions. This approach enhances the model's capacity to understand and generate data. We utilize sinusoidal positional encoding to extract features from the time steps. The periodic nature of this function captures relative relationships across different time steps, allowing the model to perceive information across a wider temporal range. We denote the resulting feature representation as $\varphi_t$.

\textbf{Backbone Training Network:}
We fuse the encoded features $\varphi_P$, $\varphi_A$, $\varphi_B$, $\varphi_L$, $\varphi_t$ with the noisy pose to obtain the overall feature $\varphi$ specific to the current task scenario. This feature is then fed into a multi-layer perceptron (MLP) structure augmented with Agent Attention, producing a $2 \times 3$ vector denoted as $\epsilon^\prime$. The vector $\epsilon^\prime$ represents the noise added by the network during the diffusion process, defined as $\epsilon^\prime = \begin{bmatrix}\epsilon_R^\prime \\\epsilon_t^\prime\end{bmatrix}$. When applying Agent Attention, we first reduce the dimensionality of the overall feature and then use it as $Q$, $K$, $V$ and $A$ in the self-attention operation. This approach enables the network to capture the interrelationships between language conditions, pose information, time step data, and point cloud features, thereby enhancing the quality and effectiveness of feature fusion. Furthermore, employing Agent Attention effectively reduces the computational complexity of Transformer-based models while retaining their strong global modeling capabilities.

\textbf{Loss:}
We adopt Smooth L1 Loss as the training loss function, which combines the advantages of both square loss and absolute loss. When the difference between the predicted value and the ground truth is small, it behaves like square loss; when the difference is large, it transitions to absolute loss. In our task, the target poses generated by the diffusion process exhibit diversity—while they may deviate significantly from the ground truth, they still represent plausible poses that allow the cup to hang securely on the rack. Therefore, using Smooth L1 Loss ensures that the loss values for such cases are not excessively large, enhancing the robustness and stability of the training process. Substituting the parameters mentioned earlier into the original Smooth L1 Loss formula, we obtain:
\begin{equation}
\small
\mathcal{L}_{SmoothL1}=\mathbb{E}_{t,T_0,\epsilon}\begin{bmatrix}\begin{cases}\frac{1}{2}\parallel\epsilon-\epsilon'(T_t,t)\parallel^2&\text{if}\parallel\epsilon-\epsilon'(T_t,t)\parallel<1\\\\\parallel\epsilon-\epsilon'(T_t,t)\parallel-\frac{1}{2}&\text{otherwise}\end{cases}\end{bmatrix}
\end{equation}
we denote the noise added to the target pose ground truth at each step as $\varepsilon_{t}$ and $\varepsilon_{R}$ (both with shape $1\times3$), corresponding to the diffusion process of the translation vector and rotation matrix, respectively. The total noise is denoted as $\varepsilon$ (with shape $2\times3$), and the noisy pose is denoted as $T_{t}$ (with homogeneous coordinates removed, shape $1\times3$. We have $\varepsilon=\begin{bmatrix}\varepsilon_R\\\varepsilon_t\end{bmatrix}$ and $T_t=[R_t,t_t^T]$.

\begin{figure*}[t!]
	\centering
	\includegraphics[width=0.9\textwidth]{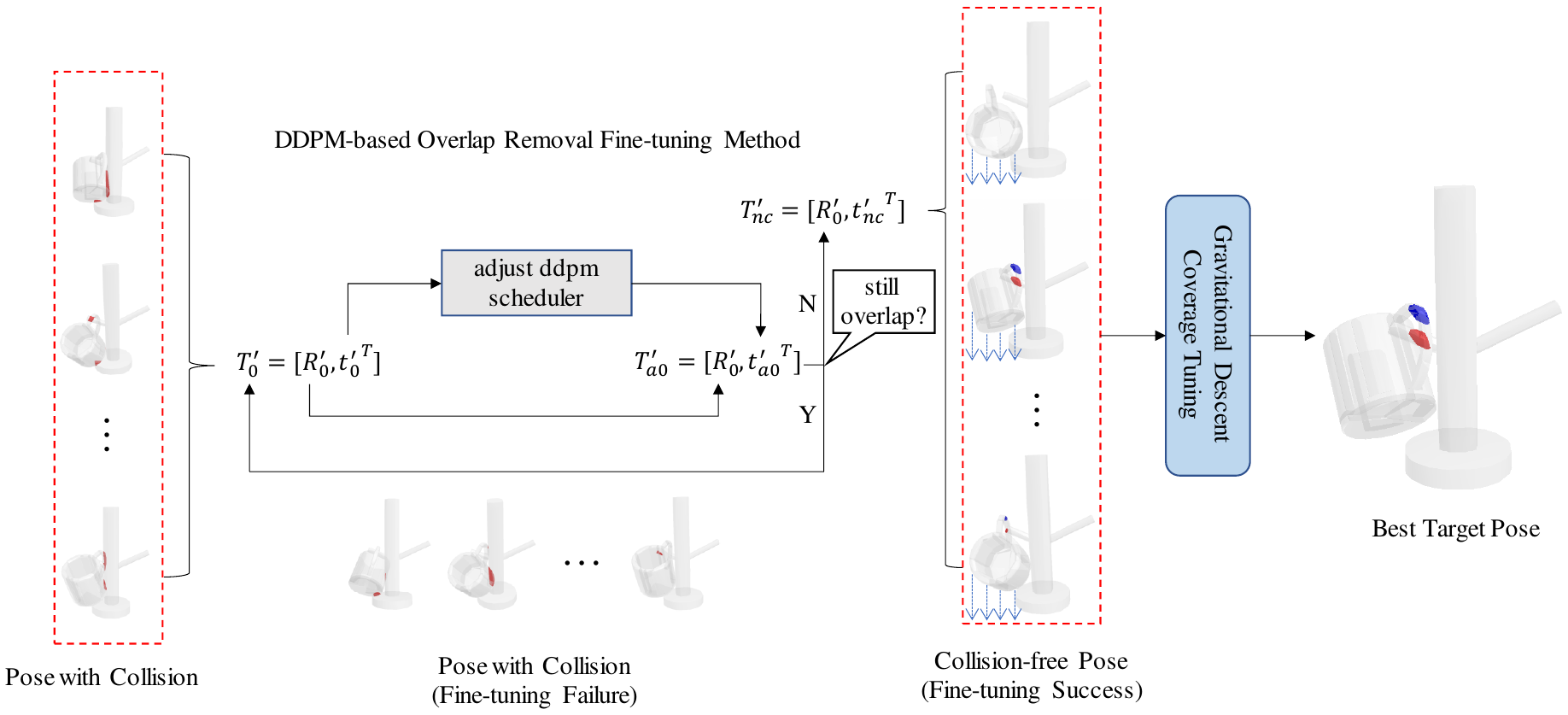} %
	\caption{The schematic diagram of the target state correction method based on the gravitational descent coverage coefficient. Here, $T_{0}^{\prime}$ represents the target pose obtained through network inference, and $T_{nc}^{\prime}$ represents the collision-free pose. We render the overlapped portion of the mug with the rack in red after the mug descends to $z_{\mathrm{opt}}$ under the collision-free pose, and render in blue the volume of this portion mapped onto the mug under $T_{nc}^{\prime}$. Finally, optimization is performed to obtain the best non-overlapping target state.}
	\label{figurelabel}
\end{figure*}

\subsection{Pose Decoupling Diffusion Module}
When performing manipulation tasks with rigid objects, as rigid bodies do not deform throughout the operation, we aim to ensure that the predicted and generated target state point clouds maintain the same shape as the initial state. To achieve this, we employ a method that generates target poses and maps them to point clouds. During the process of adding noise to poses, the rotation matrix must satisfy strict geometric constraints. Adding Gaussian noise directly to the pose matrix would disrupt the orthogonality of the rotation matrix, preventing the object from preserving its original shape. Given that the translation vector and the rotation matrix in the pose matrix are independent, we propose a Pose-Decoupled Diffusion Module that separates the pose matrix into translation and rotation components, which are then processed independently during the diffusion generation. The components are subsequently recombined to form the target pose matrix, as illustrated in the accompanying figure. We employ the normal distribution and isotropic Gaussian distribution to perform diffusion generation for the translation vector and rotation matrix, respectively. Substituting the network variables into the formulas introduced in the previous section, we define the diffusion and reverse diffusion processes of the pose decoupling diffusion module for rigid object poses, as shown in Equations (13) and (14):
\begin{equation}
\small
q(T_t\mid T_0)=\begin{cases}N\left(t_t;\sqrt{\bar{\alpha}_t}t_0,(1-\bar{\alpha}_t) I\right), &\text{for $t$}\\\mathcal{IG}_{SO(3)}\left(\lambda\left(\sqrt{\bar{\alpha}_t},R_0\right),(1-\bar{\alpha}_t)\right), &\text{for $R$}\end{cases}
\end{equation}
\begin{equation}
p(T_{t-1}\mid T_t)=\begin{cases}N\left(t_{t-1};\mu\left(t_t,t\right),\Sigma\left(t_t,t\right)\right), &\text{for $t$}\\\mathcal{IG}_{\mathcal{SO}(3)}\left(\widetilde{\mu}\left(R_tR_0\right),\widetilde{\beta}_t\right), &\text{for $R$}\end{cases}
\end{equation}
where $t_0$ and $R_0$ represent the translation vector and rotation matrix of the initial state pose, respectively, $t_t$ and $R_t$ represent the translation vector and rotation matrix at the noise addition step $t$, respectively.

\subsection{Pose Decoupling Diffusion Module}
For robotic manipulation tasks with stringent requirements for object target states, such as hanging a mug, the predicted poses generated by the network often contain errors to some extent. These errors frequently result in overlaps or collisions between the mug and the rack, adversely affecting subsequent operations. To address this issue, we propose a target state correction method based on the Gravitational Descent Coverage (GDC) coefficient. By making subtle adjustments to the target pose, this method eliminates model overlaps, as illustrated in the figure.


Specifically, experimental testing reveals that collisions in the model are primarily caused by the translational vector of the mug, while the rotational matrix has negligible impact. Therefore, similar to the noise injection process in previous networks, we decompose the predicted target pose into rotational and translational components and apply DDPM-based noise injection exclusively to the translational component. We term this process adjust ddpm scheduler. Given the relatively small pose prediction errors, we set the initial noise coefficient of the DDPM scheduler to $\beta_{start}=0.00003$ and the timestep to 0, as detailed in Figure 6. After applying the de-overlapping post-processing module, we obtain a collision-free target pose, denoted as $T_{a0}^{\prime}=[R_{a0}^{\prime},{t_{a0}^{\prime}}^{T}]$. This pose can be directly fed into downstream robotic motion planning, significantly improving the task success rate.

Specifically, during the fine-tuning process, there may be instances where the objects do not collide but fail to complete the intended task. Additionally, the feasible states obtained through fine-tuning can vary, requiring us to select the most suitable state as the input for downstream robotic motion planning to enhance task success rates. To filter out unsuitable poses and prioritize the fine-tuned target states, we adopt an evaluation method based on the GDC coefficient.In detail, the fine-tuned target pose is mapped to the model and projected along the $z$-axis. To simulate the scenario where the mug falls downward due to gravity when the gripper is released, the mug model is "descended" within a certain range in the $z$-direction. During this descent, we compute the maximum ratio of the overlapping volume between the mug and the rack to the total volume of the rack. This ratio is referred to as the GDC coefficient, denoted as $C_{\mathrm{GDC}}$, and it is described by the following formula:
\begin{equation}
C_{\mathrm{GDC}}(z_{\mathrm{opt}})=\max_{z_i\in[0,z_{\mathrm{max}}]}\{\frac{\mathrm{Vol}(R\cap M(z_i))}{\mathrm{Vol}(R)}\}
\end{equation}
where $z_{\mathrm{opt}}$ denotes the downward displacement at which the overlapping volume reaches its maximum, $z_i\in[0,z_{\mathrm{max}}]$ represents the range of descent, $R$ represents the rack model, $M(z_i)$ refers to the mug model at a descending distance , and $\mathrm{Vol}$ represents the volume of a model. By comparing the $C_{\mathrm{GDC}}$ values of different fine-tuned target states, we can avoid cases where there is no overlap but the hanging task fails. This comparison allows us to prioritize and select the target state with the highest $C_{\mathrm{GDC}}$ for downstream robotic motion planning, thereby enhancing the robustness and accuracy of the operational task.

\section{Experiments}

To evaluate our method, we conducted experiments based on the placement tasks in NDF, across three task environments: single-mode (a single hook), multi-mode (multiple hooks), and designated mode (using language conditions to specify one of the multiple hooks). In each experiment, we provided a randomly initialized pose for the mug rack and a random pose for the mug, using a network to predict the pose of the mug when successfully hung on the rack. The experiments were conducted using five different mugs and five different racks, with each rack containing ten hooks. In prior work, such as RELDIST, the success criteria accounted for overlapping regions between objects by defining three thresholds: $<$1 cm, $<$3 cm, and $\infty$, and the success rate was calculated under each threshold. However, we argue that if the predicted pose is to be directly used in downstream motion planning for robotic arms, it is crucial to ensure that there is no overlap between the mug and the rack. Overlapping poses would significantly disrupt the experiment and could introduce safety risks, such as a robotic arm forcibly colliding the mug with the rack to achieve a predicted overlapping pose. Therefore, we adjusted the success criteria as follows: the hook must pass through the mug handle, and there must be no overlap between the mug and the rack. For post-processing, we set a maximum of 100 iterations. If overlapping remains after the iterations, the hanging task is deemed a failure. We conducted 50 trials for each combination of mugs and hooks under the three experimental settings: single-mode, multi-mode, and designated mode. We record success rates and errors in each case. We divide the error between the predicted pose and the target pose into two components: rotation and translation. The rotation error is measured by the geodesic rotation distance $\epsilon_R$, and the translation error is represented by the $L_2$ distance $\epsilon_t$. Additionally, we record the average fine-tuning count $T_{avg}$, the success rate without fine-tuning (i.e., when the predicted mug target state is correct and there is no overlap) $SR_{nt}$, and the overall success rate $SR_{total}$.

\subsection{Dataset Collection}
To obtain the ground-truth target poses for mugs hanging on racks, we utilized the Robosuite [22] simulation environment, which incorporates a realistic physics engine, to collect data due to the physical contact between mugs and racks in their ground-truth poses. For the mugs, we selected 10 distinct mug models from the ShapeNet [23] dataset. For the racks, we constructed five different rack models, each designed with two distinct hooks to differentiate them, resulting in a total of 10 unique hooks. These hooks correspond to 10 different language conditions, as shown in Figure 4. To validate the diversity of feasible target states generated by the model, our dataset includes five different pose demonstrations for each mug on each hook. Each demonstration contains the current mug point cloud, the current rack point cloud, the ground-truth target pose, the language condition, and its encoded representation. The language conditions consist of 10 phrases: "place the mug on the (longer/shorter/higher/lower/horizontal/tilted/rectangular/
cylindrical/curved/straight) rack." In addition to validating the generalization of the seen mug in different initial states, we selected five previously unseen mug models from the ShapeNet dataset to test the generalization capability of our method, as illustrated in Figure 4.

   \begin{figure}[thpb]
      \centering
      \includegraphics[width=\columnwidth]{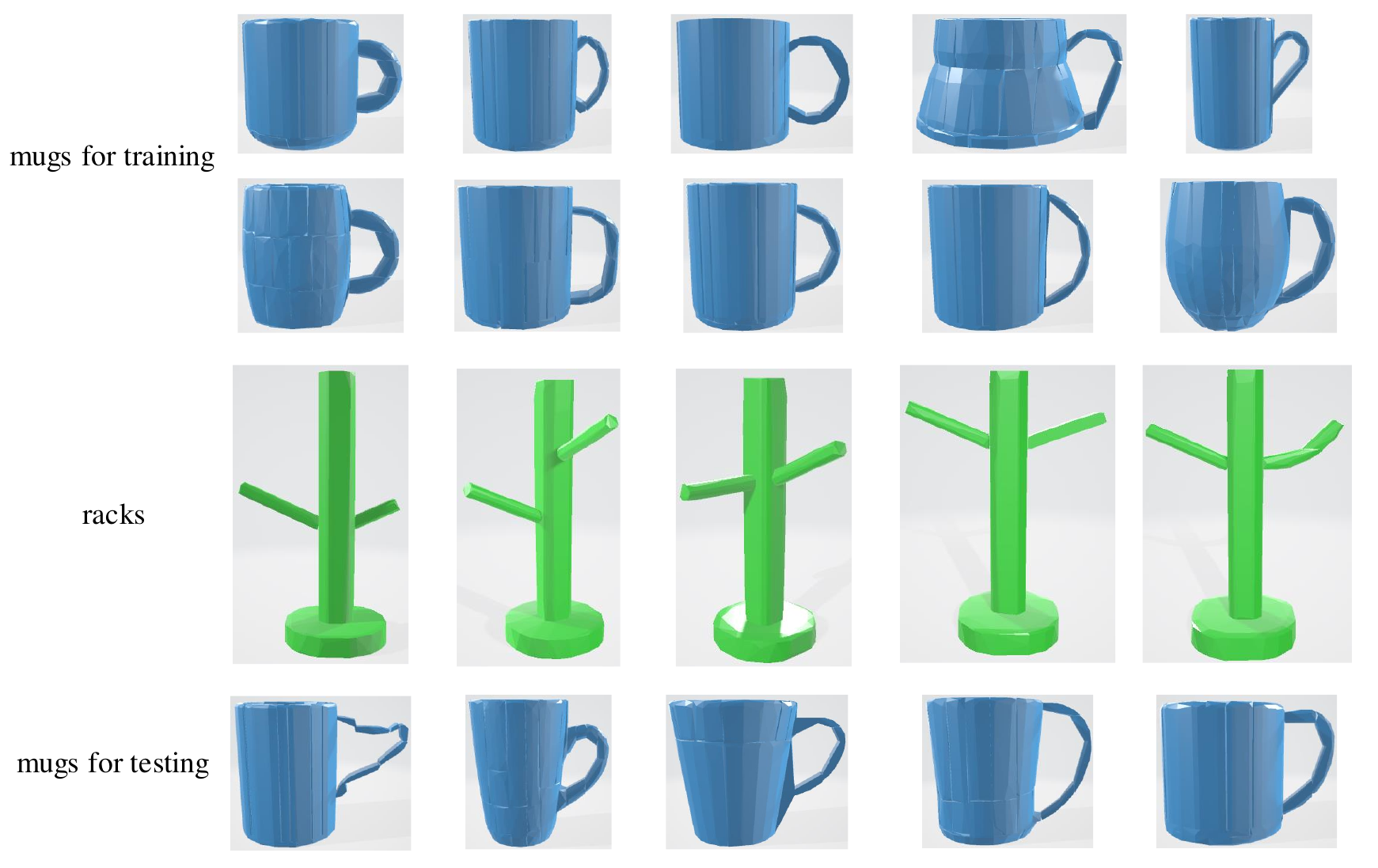}
      \caption{The training mug models (top), the training rack models (middle), and the testing mug models (bottom).}
      \label{figurelabel}
   \end{figure}

\subsection{Method Characteristics}
\textbf{Multi-Modal Distribution.}
Compared to the baseline method, Tax-Pose, the primary advantage of our approach lies in LHGD’s capability to handle multi-modal distribution tasks. Specifically, we designed a multi-modal distribution task as follows: since each mug rack has two hooks located at different positions, the task is considered successful if, starting from any initial state of the mug, a valid target state is generated where the mug is placed on either hook. Additionally, LHGD is adaptable to two different types of input: one without language-conditioned input and another with language-conditioned input.For the multi-modal distribution task with language-conditioned input, to ensure that the generated target states are randomly distributed across the two hooks, we assign identical language conditions to both hooks. Specifically, we use the instruction place the mug on the arbitrary rack, indicating that the mug can be hung on either hook in this task. The results of the two input methods and corresponding comparisons are presented in Table $\textnormal{\uppercase\expandafter{\romannumeral1}}$.
\begin{table}[htbp]
  \centering
  \caption{Multi-Modal Distribution}
    \begin{tabular}{cccccr}
    \toprule
    \multicolumn{3}{c}{}  & \multicolumn{2}{c}{seen mugs} & \multicolumn{1}{c}{unseen mugs} \\
    \midrule
    \multicolumn{3}{c}{Tax-Pose} & \multicolumn{2}{c}{0} &  \multicolumn{1}{c}{0}     \\
    \midrule
    \multicolumn{3}{c}{LHGD(without language condition)} & \multicolumn{2}{c}{90.4\%} &    \multicolumn{1}{c}{86\%}   \\
    \midrule
    \multicolumn{3}{c}{LHGD(with language condition)} & \multicolumn{2}{c}{85.6\%} & \multicolumn{1}{c}{84.8\%}     \\
    \bottomrule
    \end{tabular}%
  \label{tab:addlabel}%
\end{table}%

Experiments show that in a multi-modal distribution, there are two potential distributions (corresponding to two hanging rods) when completing the task. As a result, the TAX-Pose method, which is suited for single-modal distributions, generates a target state that lies between the two potential targets, making it unable to complete the task. The diffusion model used in our method generates data through a stepwise reverse sampling process. It can learn different peak regions based on the characteristics of the training data distribution during the generation process and correctly reproduce them. Therefore, it performs well in multi-modal distribution tasks and maintains a more even distribution. Figure 5 shows the visualization results of multi-modal distributions for different methods.
   \begin{figure}[thpb]
      \centering
      \includegraphics[width=\columnwidth]{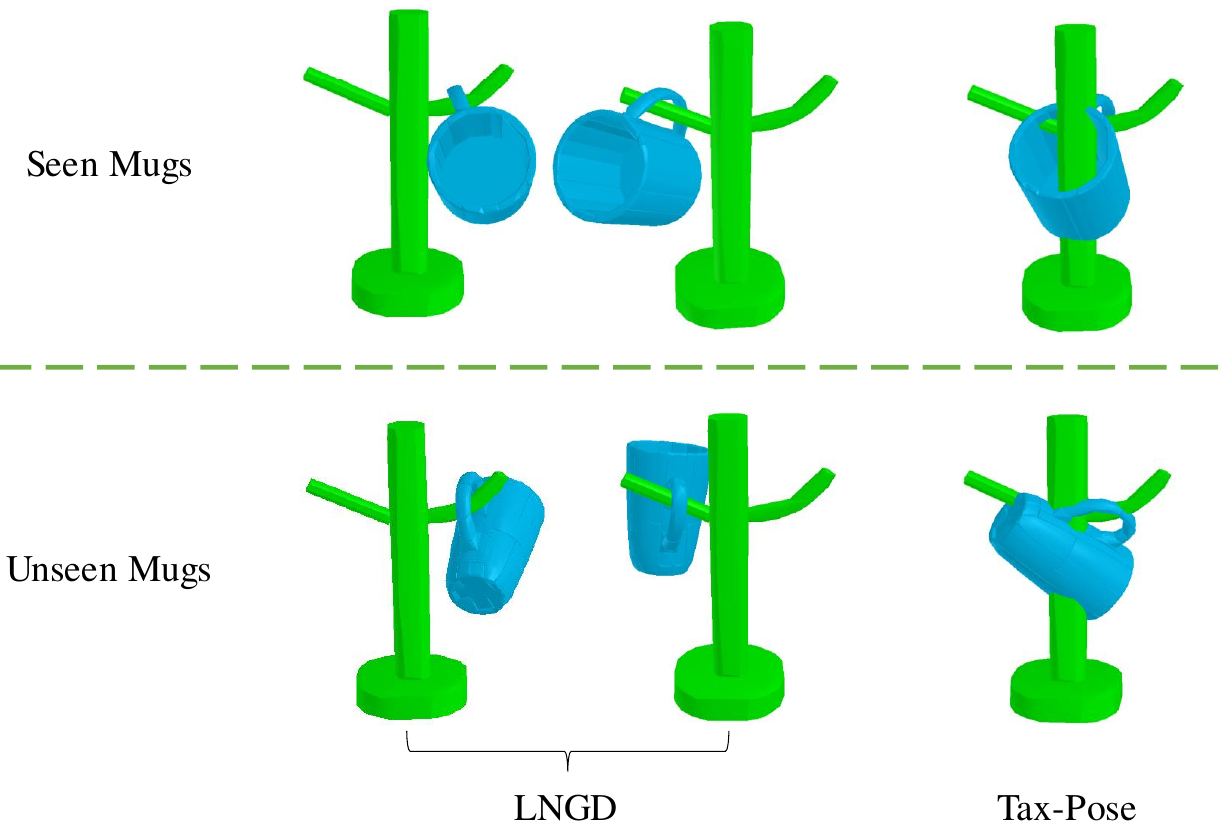}
      \caption{Visualization results of different methods on multi-modal distribution tasks.}
      \label{figurelabel}
   \end{figure}

\textbf{Single-mode Distribution}
Our method not only handles the more complex multi-modal distribution but also successfully completes simpler single-mode distributions, as seen in Tax-Pose. Specifically, it generates the target state of the mug when the task is completed on a shelf with only one hanging rod. Through experiments, we found that during data collection, there could be multiple possible final target states for the mug (e.g., the mug's opening facing either the left or right when hung). If these variations are included in the dataset and Tax-Pose is trained on it, the same compromise issue as in the multi-modal distribution arises. This requires that each mug be collected with a consistent orientation for every different hanging rod, which imposes significant constraints on data collection and is clearly unreasonable. In contrast, our method does not require such constraints when collecting data. To validate this characteristic, we constructed two datasets: Dataset 1, where the target orientation is uniform, and Dataset 2, which contains multiple target orientations. We tested Tax-Pose and LHGD on these two datasets, and the results are shown in Table $\textnormal{\uppercase\expandafter{\romannumeral2}}$, with the visualization results provided in Figure 6. As shown in the table, our method outperforms Tax-Pose on both datasets. Moreover, through target state correction, our accuracy is significantly improved.
\begin{table}[htbp]
  \centering
  \caption{Single-mode Distribution}
    \begin{tabular}{cccccrrr}
    \toprule
    \multicolumn{3}{c}{}  & \multicolumn{2}{c}{$SR_{nt}$} & \multicolumn{1}{c}{$SR_{total}$} & \multicolumn{1}{c}{$\epsilon_R$} & \multicolumn{1}{c}{$\epsilon_t$} \\
    \midrule
    \multicolumn{3}{c}{Tax-Pose(dataset1)} & \multicolumn{2}{c}{2.8\%} &   \multicolumn{1}{c}{/}    &      \multicolumn{1}{c}{56.84} & \multicolumn{1}{c}{0.0093} \\
    \midrule
    \multicolumn{3}{c}{LHGD(dataset1)} & \multicolumn{2}{c}{29.2\%} &  \multicolumn{1}{c}{94.0\%}     &    \multicolumn{1}{c}{23.36}   & \multicolumn{1}{c}{0.0072} \\
    \midrule
    \multicolumn{3}{c}{Tax-Pose(dataset2)} & \multicolumn{2}{c}{31.2\%} &   \multicolumn{1}{c}{/}    &    \multicolumn{1}{c}{16.94}   & \multicolumn{1}{c}{0.0080} \\
    \midrule
    \multicolumn{3}{c}{LHGD(dataset2)} & \multicolumn{2}{c}{33.6\%} &   \multicolumn{1}{c}{92.8\%}    &   \multicolumn{1}{c}{22.12}    & \multicolumn{1}{c}{0.0094} \\
    \bottomrule
    \end{tabular}%
  \label{tab:addlabel}%
\end{table}%

   \begin{figure}[thpb]
      \centering
      \includegraphics[width=\columnwidth]{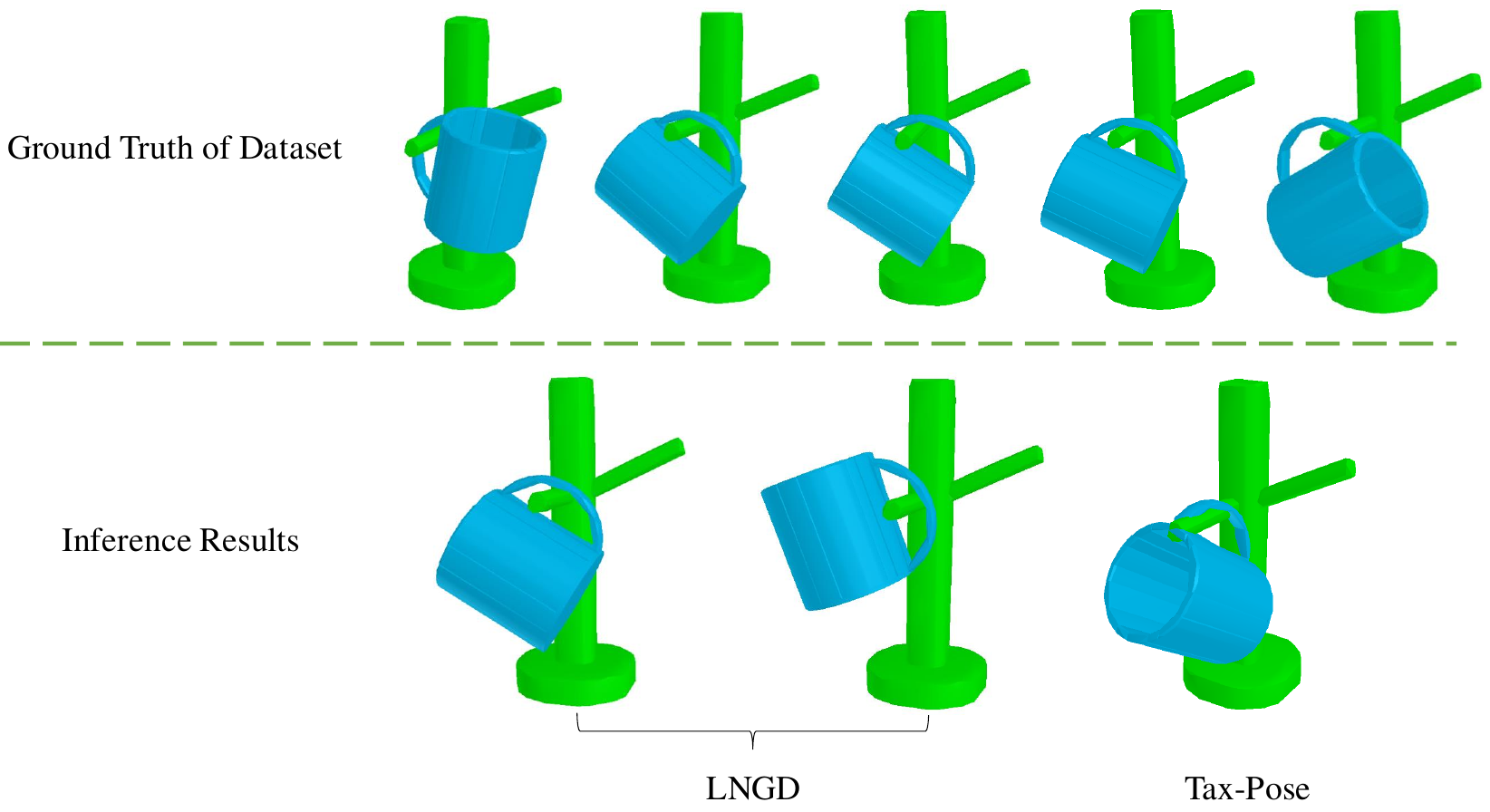}
      \caption{The inference results after training both methods on Dataset 2. LHGD is not affected by the target state distribution of the dataset and is able to generalize to unseen target states, while Tax-Pose is significantly influenced by the dataset, resulting in poorer inference results.}
      \label{figurelabel}
   \end{figure}

\textbf{Language Sensitivity.}
Our method uses language to control the generation of target states, requiring high sensitivity to language in order to generate the correct target states corresponding to different language conditions. Moreover, for erroneous language instructions, the method should be able to distinguish them and output the incorrect target states, ensuring a one-to-one correspondence between input and output, thus reflecting the method's sensitivity to language conditions. We use CLIP as the language condition encoder precisely because it can capture subtle semantic differences when representing language. We conduct tests using a mug on a shelf that differentiates between high and low positions, and the experimental results are shown in Table $\textnormal{\uppercase\expandafter{\romannumeral3}}$.
\begin{table}[htbp]
  \centering
  \caption{Language Sensitivity}
    \begin{tabular}{cccccrr}
    \toprule
    \multicolumn{3}{c}{}  & \multicolumn{2}{c}{$SR_{total}$} & \multicolumn{1}{c}{$T_{avg}$} & \multicolumn{1}{c}{$SR_{nt}$} \\
    \midrule
    \multicolumn{3}{c}{LHGD(place the mug on the higher rack)} & \multicolumn{2}{c}{99.2\%} &   5.85    & 51.6\% \\
    \midrule
    \multicolumn{3}{c}{LHGD(place the mug on the lower rack)} & \multicolumn{2}{c}{97.6\%} &   3.19   &  60.8\%\\
    \midrule
    \multicolumn{3}{c}{LHGD(place the mug on the red rack)} & \multicolumn{2}{c}{0\%} &   \multicolumn{1}{c}{/}   &  \multicolumn{1}{c}{0\%}\\
    \midrule
    \multicolumn{3}{c}{LHGD(place the mug on the red rack)} & \multicolumn{2}{c}{0\%} &   \multicolumn{1}{c}{/}   &  \multicolumn{1}{c}{0\%}\\
    \midrule
    \multicolumn{3}{c}{LHGD(place the mug on the red rack)} & \multicolumn{2}{c}{0\%} &   \multicolumn{1}{c}{/}   &  \multicolumn{1}{c}{0\%}\\
    \bottomrule
    \end{tabular}%
  \label{tab:addlabel}%
\end{table}%

To demonstrate the linguistic sensitivity of our method, we also tested the generated target states when inputting correct and incorrect language conditions, with the visualization results shown in the figure 7. The results indicate that our method can clearly distinguish subtle variations in input language conditions, where correct language conditions lead to better target states for the mug, while incorrect language conditions cause a significant deviation from the expected result, reflecting good linguistic sensitivity.

   \begin{figure}[thpb]
      \centering
      \includegraphics[width=\columnwidth]{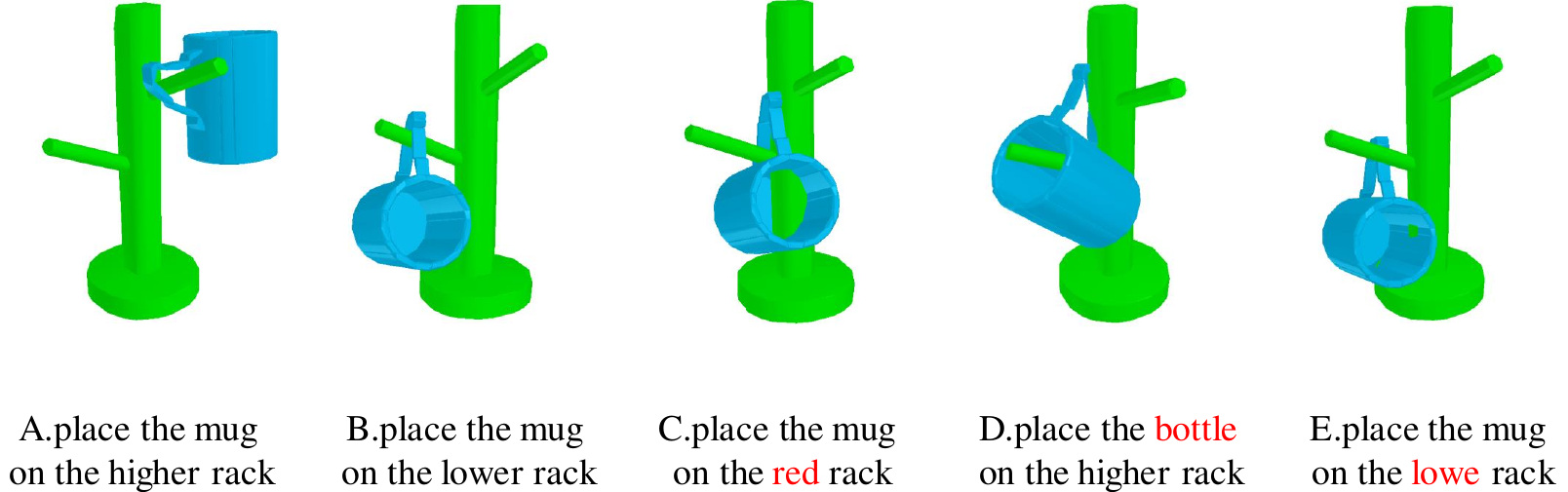}
      \caption{The visualization results of linguistic sensitivity, where A and B represent scenarios with suspension based on correct language conditions; C and D represent incorrect language conditions with one word replaced, which still align with common sense; E represents an incorrect language condition with a spelling error.}
      \label{figurelabel}
   \end{figure}

\textbf{Generative Diversity.}
The operation task of placing the mug on the rack has multiple possible target states. Our method, LHGD, is capable of fitting the target distribution while preserving the local characteristics of the data. It outputs multiple possible target states based on conditions, which facilitates the subsequent optimization of these outputs through target state correction methods. We believe that, while ensuring successful placement, if there is a significant deviation in the values of $\epsilon_R$ and $\epsilon_t$, it can be considered that the model has generalized to target poses not present in the ground truth, indicating good diversity in the generated target states. To visualize this diversity, we show different output results for both seen and unseen mugs placed on the language-specified long rod in figure 8. For the seen mugs, we calculate their $\epsilon_R$ and $\epsilon_t$ to intuitively demonstrate the diversity of the generated target states.

   \begin{figure}[thpb]
      \centering
      \includegraphics[width=\columnwidth]{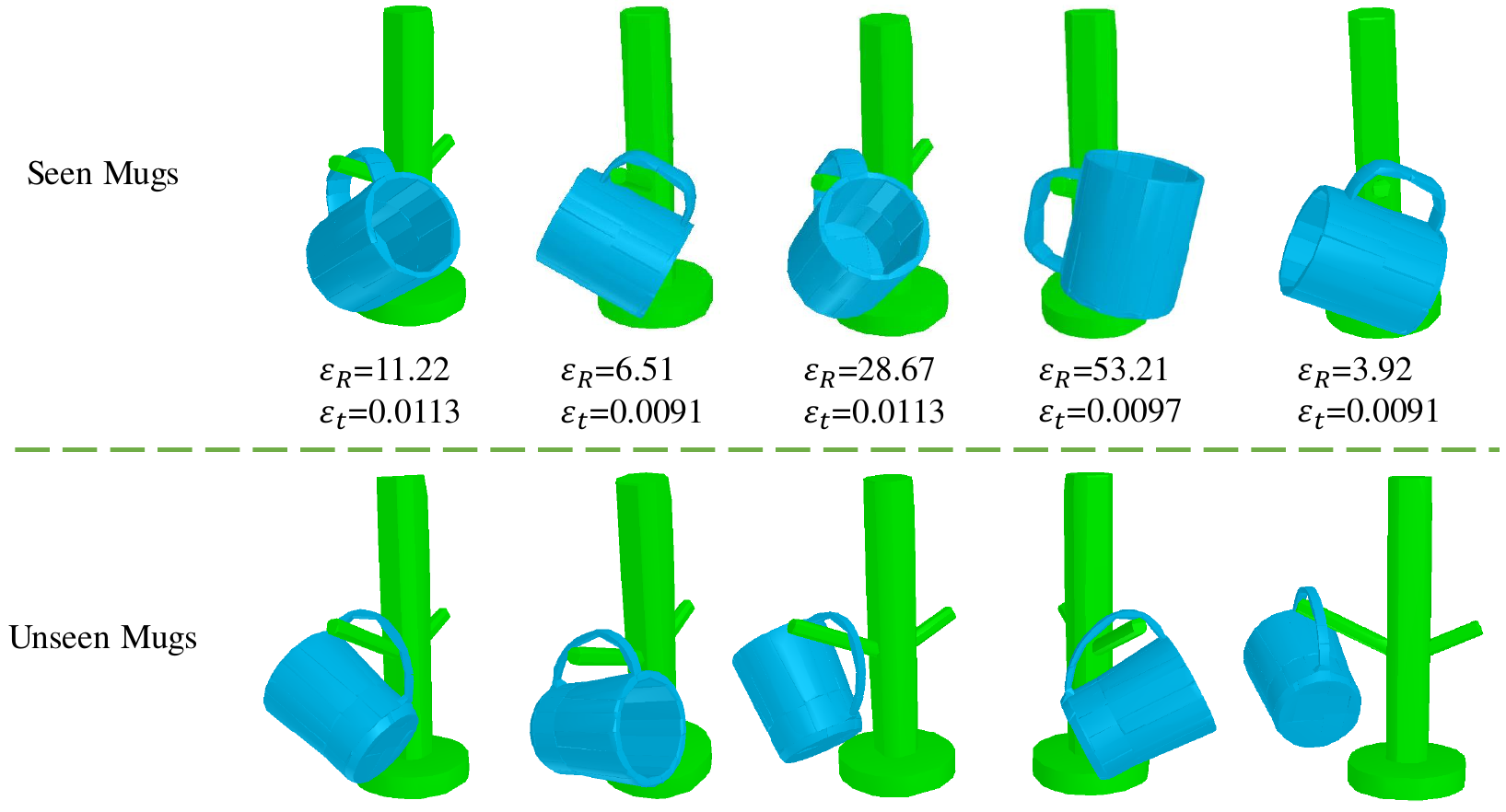}
      \caption{Visualization of the Diversity of Generated Target States for Seen and Unseen Mugs.}
      \label{figurelabel}
   \end{figure}

\textbf{The Impact of Correcting Time Steps.}
In the DDPM scheduler, timesteps represent the time steps of the process where noise is gradually added to the samples. The different values of timesteps have a significant impact on our target state correction module. The larger the value, the greater the amount of noise added, which leads to the corrected target state being further from the predicted value. In our target state correction process, we only aim to perform the forward diffusion process, so timesteps can be treated as a hyperparameter for ablation evaluation. Since the target state obtained through the model is very close to the desired collision-free state, we hope that the noise added can diffuse the predicted value to the collision-free state, while also minimizing the number of processing steps to improve efficiency. Based on this, we conducted experiments to find the optimal noise addition timestep $t$, setting this hyperparameter to $t=1,3,6,10$ and testing on an unseen mug. The results are shown in Figure 9.

\begin{figure}
\begin{center}
\subfigure[Success Rate]{
\includegraphics[width=\columnwidth]{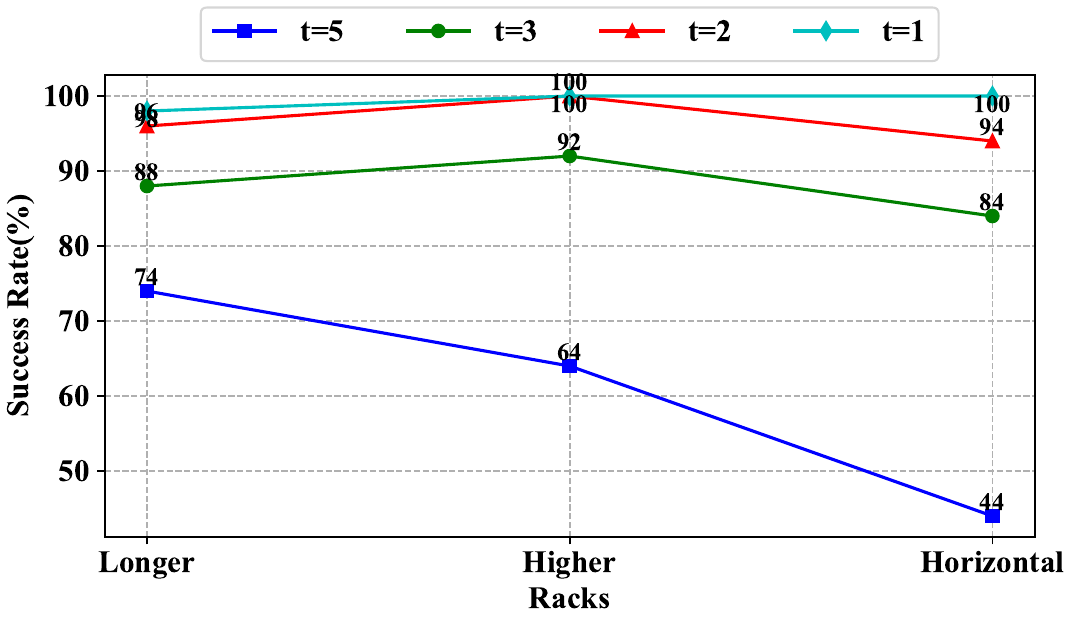}
}
\subfigure[Average Fine-Tuning Steps]{
\includegraphics[width=\columnwidth]{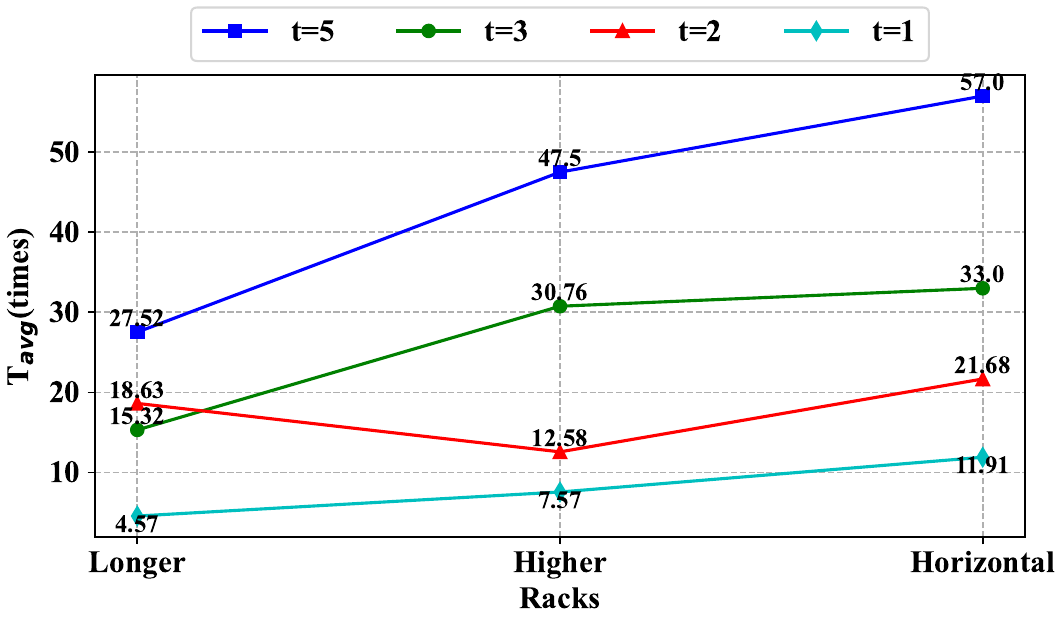}
}
\caption{The two plots on the left and right represent the success rate and the number of fine-tuning steps for different timesteps on several types of hangers. It can be observed that the success rate is higher at timesteps $t=1$ and $t=2$, while the success rate is lower at $t=3$ and $t=5$. Furthermore, when the success rates are similar, fewer fine-tuning steps are required at $t=1$. Therefore, setting the timesteps to $t=1$ is the optimal solution, as it balances both the success rate and task completion efficiency.}
\end{center}
\end{figure}

\section{CONCLUSION}

We propose LHGD, a model that utilizes conditional diffusion models to predict target states. Through multiple sets of experiments, we have demonstrated that LHGD performs well in robotic manipulation tasks, including single-mode (single hook), multi-mode (multiple hooks), and language-conditioned mode distribution (specific hook). To address the potential overlap issues between the predicted cup and rack, we designed a target state correction method based on the gravitational descent coverage coefficient, which effectively resolves the overlap problem and achieves the best results across various tasks. Although LHGD performs well in tasks with language-conditioned constraints and significantly reduces overlap issues, there are still some limitations to our method:

\begin{itemize}
\item Model Dependency. While our post-processing overlap removal is quite accurate, it relies on object models. For point clouds of unseen objects collected in the space, the method may not perform well.
\item Language Condition Generalization. Although our model generalizes to unseen mugs to some extent, when testing on racks with completely new and unseen language conditions (e.g., racks with different colored hooks), our model may not capture the semantics as effectively as larger models, leading to suboptimal performance in completing the task.
\end{itemize}

\addtolength{\textheight}{-12cm}   









\end{document}